\documentclass[letterpaper]{article} 
\usepackage{aaai23}  
\usepackage{times}  
\usepackage{helvet}  
\usepackage{courier}  
\usepackage[hyphens]{url}  
\usepackage{graphicx} 
\usepackage{natbib}  
\usepackage{caption} 
\usepackage{subcaption}
\usepackage{pgfplots, pgfplotstable}

\usepackage{newfloat}
\usepackage{listings}
\DeclareCaptionStyle{ruled}{labelfont=normalfont,labelsep=colon,strut=off} 
\newcommand{\calypso}{\textsc{Calypso}}

\title{\calypso{}: LLMs as Dungeon Masters' Assistants}
\author {
    Andrew Zhu$^1$,
    Lara J. Martin$^2$\footnote{Work done while at the University of Pennsylvania.},
    Andrew Head$^1$,
    Chris Callison-Burch$^1$
}
\affiliations {
    $^1$University of Pennsylvania\\
    $^2$University of Maryland, Baltimore County\\
    \texttt{\{andrz, head, ccb\}@seas.upenn.edu, laramar@umbc.edu}
}

\begin{document}

\maketitle

\begin{abstract}
    The role of a Dungeon Master, or DM, in the game Dungeons \& Dragons is to perform multiple tasks simultaneously.
    The DM must digest information about the game setting and monsters, synthesize scenes to present to other players, and respond to the players' interactions with the scene.
    Doing all of these tasks while maintaining consistency within the narrative and story world is no small feat of human cognition, making the task tiring and unapproachable to new players.
    Large language models (LLMs) like GPT-3 and ChatGPT have shown remarkable abilities to generate coherent natural language text.
    In this paper, we conduct a formative evaluation with DMs to establish the use cases of LLMs in D\&D and tabletop gaming generally. We introduce \calypso{}, a system of LLM-powered interfaces that support DMs with information and inspiration specific to their own scenario. 
    \calypso{} distills game context into bite-sized prose and helps brainstorm ideas without distracting the DM from the game.
    When given access to \calypso{}, DMs reported that it generated high-fidelity text suitable for direct presentation to players, and low-fidelity ideas that the DM could develop further while maintaining their creative agency.
    We see \calypso{} as exemplifying a paradigm of AI-augmented tools that provide synchronous creative assistance within established game worlds, and tabletop gaming more broadly.
    
\end{abstract}

\section{Introduction}



Dungeons \& Dragons (D\&D) \cite{Gygax1974} is a tabletop role-playing game (TTRPG)---a collaborative storytelling game where a group of players each create and play as their own character, exploring a world created by and challenges set by another player known as the Dungeon Master (DM). It is the DM's role to play the non-player characters and monsters, and to write the overarching plot of the game.

\begin{figure}[ht]
    \centering
    \includegraphics[width=\columnwidth]{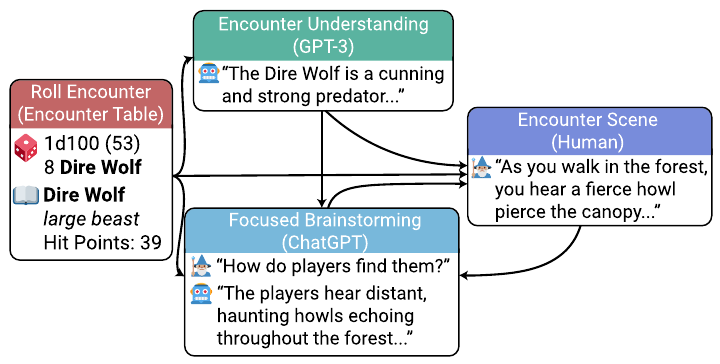}
    \caption{After rolling a random encounter (red), DMs can use LLMs with \calypso{} to help generate an encounter scene and digest information about monsters. \calypso{} can present monster information concisely (green) and brainstorm conversationally (purple) to help build a compelling narrative to present to players (purple).}
    \label{fig:front}
\end{figure}

As a co-creative storytelling game, Dungeons \& Dragons presents multiple unique challenges for AI systems aiming to interact with it intelligently. Over the course of a game, which is played out across multiple sessions spanning a long duration of time (often multiple months to years), the DM and the other players work together to produce a narrative grounded in commonsense reasoning and thematic consistency \cite{ammanabrolu_bringing_2020, bergstrom_framing_2011}. As the group plays for longer, the players define more of the world and ad-hoc rules for interacting with it \cite{van_velsen_table-top_2009}. In order to make in-character decisions, each individual player must maintain a personal understanding of the game world which they build from the game history \cite{Martin2018} while keeping track of what information other players and their characters know \cite{Zhou2022}.


By using an AI co-DM tool, human DMs can devote more mental energy to cognitively demanding tasks of being a DM, such as improvising dialog of NPCs (non-player characters) or repairing the script of their planned campaign.
Furthermore, an AI co-DM would drastically reduce the barrier of entry into DMing.
Therefore, an AI co-DM tool would be invaluable to the D\&D community.

An effective AI co-DM tool should not only produce coherent and compelling natural language output for a DM to effectively use for inspiration but also account for an immense amount of background context and requirements for internal consistency---both within D\&D rules and within a given scenario or campaign. 
Large language models (LLMs), such as GPT-3 \cite{GPT3} and ChatGPT \cite{ChatGPT}, have shown impressive abilities to generate coherent text.
Some \cite{Callison-Burch2022, FIREBALL} have even applied LLMs to the problem of D\&D dialog and narrative by finetuning the models with structured information.
Whereas these works used structured information scraped from user data to fine-tune a single model, we use existing data in D\&D source books to improve generation using zero-shot prompting with multiple models.


In this paper, we present a study in which we created a LLM-augmented tool to assist DMs in playing D\&D. We employed the following methods:
\begin{enumerate}
    \item We interviewed DMs to understand how they digest game information and learn design motivations for AI assistants in the domain.
    \item We created a gameplay setting that allowed us to study D\&D gameplay on a larger scale than other recent works and invited 71 players to participate.
    \item We created a system of three LLM-powered interfaces, which we call \calypso{} (\textbf{C}ollaborative \textbf{A}ssistant for \textbf{L}ore and \textbf{Y}ielding \textbf{P}lot \textbf{S}ynthesis \textbf{O}bjectives), that DMs and players could use as they played D\&D, and studied the ways in which DMs and players incorporated them into their creative process over four months using established HCI methods.
\end{enumerate}

We show that language models are capable ``co-DMs" -- not a player in the same way that the human players and DM are, but still a synchronous agent that acts as a guide for the human DM. We provide insights into how TTRPG players actually want to use these tools and present validated solutions that can extend beyond the D\&D domain. Our study shows that a system designed with these motivations in mind saw consistent prolonged usage among a community of creative writers.




\section{Background and Related Work}

\subsection{Dungeons \& Dragons in the Time of COVID}
\label{sec:dnd}
Traditionally, Dungeons \& Dragons is played in person. Players use physical character sheets and monster stats referenced from books containing hundreds of prewritten ``stat blocks" (as pictured in Figure \ref{fig:summary_info}) \cite{DnD_MM}. DMs have the option to create a world of their own to play in (also sometimes called ``homebrewing" a setting) or to set their game in a professionally written ``module": a book containing a detailed outline of an adventure, including the setting, non-player characters, predesigned challenges and monster encounters, and lore. Previous works have explored methods of how to present information in these existing settings more clearly to DMs, such as through a computer-generated adventure flowchart \cite{acharya_interviews_2021} or recommender systems for relevant entities in a scene \cite{perez_synset-based_2021}.

Since the beginning of the COVID-19 pandemic, there has been a shift towards playing D\&D online \cite{yuan_tabletop_2021}. Rather than using physical character sheets and reference books while playing in person, a large number of groups instead play virtually using tools like \citet{DnDBeyond} for virtual character sheets and reference books, Discord for messaging, virtual tabletops like Foundry \cite{FoundryVTT} to simulate maps, and game state trackers like Avrae \cite{Avrae} to track character and monster stats. For inspiration and immersion, DMs also use online tools like \citet{dScryb}, which provides prewritten text, Tabletop Audio \cite{TabletopAudio}, which provides soundboards and soundscapes, and random tables published in D\&D source books \cite{DnD_DMG}, which provide a prewritten set of options, for specific scenarios (e.g. encountering a dragon).

\subsection{Large Language Models and D\&D}
Large language models (LLMs) are a recent development in the area of Natural Language Processing that have demonstrated emergent capabilities of understanding users' input and replying directly in the user's language (c.f. a machine language). A neural architecture based on the Transformer \cite{transformer}, they are capable of learning user-defined tasks with no additional training (``few-shot" or ``in-context" learning) and referencing concepts defined in their large training corpus \cite{GPT3}.

Although there has been some work looking at playing Dungeons \& Dragons using earlier neural language models \cite{Louis2018,Martin2018,Rameshkumar2020}, the introduction of LLMs has created a renewed interest in researching tabletop gaming.
\citet{Callison-Burch2022} frame D\&D as a dialogue challenge and examine whether LLMs are capable of predicting a player's next utterance based on the conversational history, finding that local game-specific state context is important for grounded narrative generation.
\citet{newman_generating_2022} use LLMs to generate novel material (namely spells) that is consistent with the style and rules of the game.
\citet{Zhou2022} create a system that models the intents of D\&D players using LLMs to inform a surrogate Theory of Mind.
\citet{FIREBALL} instrument a game state tracker to provide concrete actor stats and combat state, finding that LLMs are capable of producing interesting roleplay in combat scenarios and predicting the action a player will take. They highlight the importance of player and DM agency in LLM-generated texts, proposing that LLMs are better suited for assistant-style use cases.
\citet{kelly_towards_2023} present a preliminary work using LLMs to identify player questions from live transcriptions of gameplay and suggest in-character responses.

\citet{santiago_iii_rolling_2023} have proposed multiple scenarios where LLMs and other generative AI models may be used to assist DMs, and discuss the ways AI may be used. In this workshop paper, they hypothesize the potential for AI to help inspire and take cognitive burden off the DM and provide brainstorming inspiration, but also weaknesses where AI may fall back onto overused tropes or underrepresent minority groups.
In this work, we explore and expand upon many of these hypotheses through interviews with DMs. We create a system where DMs can fluently incorporate a LLM into their creative process and run a broad study on its use and failure cases.

LLMs have been explored as a writing assistant in other modalities as well, using various methods to assist in collaboratively building a narrative. These works have examined the use of conversational agents \cite{coenen_wordcraft_2021, ippolito_creative_2022}, writing in established settings \cite{akoury_storium_2020}, and other human-in-the-loop methods \cite{chung_talebrush_2022, schoenau-fog_creative_2015, nack_design_2016, calderwood_how_2020, Yang2022, kreminski_loose_2022}. There has also been work proposing LLMs for multimodal co-creative frameworks \cite{lin_creative_2022}. Overall, these techniques differ from D\&D and other TTRPGs in that they primarily focus on a single writer/creator interacting with the system, rather than the multi-player experience in TTRPGs where all players directly interact with the story.

To our knowledge, our work is the first to examine concrete implementations of multiple unique interaction modalities in and outside of combat scenarios and the ways D\&D players interact with language models on this scale.

\section{Design Motivation}
To better understand the friction DMs face in looking up reference material midgame, we conducted interviews and ran workshop sessions with seven DMs (referred to as D1-7 below) from a wide range of backgrounds before creating our system. Participants ranged from 1 to 39 years of experience playing D\&D (various editions). In these sessions, we asked DMs how they approached improvising encounters -- i.e., to run random encounters that are generated on the fly (usually by rolling on an encounter table). In random encounters, DMs do not have time to research the monster's stats and lore beforehand and think of backstories as to why the monster ended up in a particular setting. 
From these interviews, we identify several ways how an AI system could be helpful to DMs:

\paragraph{Inspiration.}
As proposed by \citet{santiago_iii_rolling_2023}, we find that DMs desired the ability to use a language model to generate the first draft of an encounter, which they could then build on top of with their own ideas (D1-3). Different DMs envisioned giving the system varying amounts of control over the narrative. D3 expressed that they would want a system to write a scene that they would then vet and choose whether to present it verbatim to their players, edit it to their liking, or use as inspiration to overcome writer's block. D1 and D2 envisioned using the system's generation verbatim to present an initial scene to players while they either read the complete text of the monster description (D2) or to reduce cognitive load (D1).

\paragraph{Strategic Copilot.}
\label{sec:strategic-copilot}
One DM mentioned that managing both narrative gameplay and tracking monster stats and mechanics overwhelmed their short-term memory, and expressed interest in a system that could aid them in making strategic decisions and acting as a high-level copilot. They expressed that the large amount of low-level management was a barrier to them running more D\&D, and that they wanted to ``feel more like an orchestra conductor over someone who's both putting down the train tracks AND fueling the train" (D4).

Another DM said that DMs often fail to take into account monsters' unique abilities and stats when running encounters, making simplifications to manage a large number of monsters. For example, a monster with very high intelligence and low dexterity attempting to move sneakily ``should know not to move and make a bunch of noise" (D6).

\paragraph{Thematic Commonsense.}
\label{sec:thematic-commonsense}
We asked DMs what parts of monsters' game statistics they found to be the most important for their understanding of how to use a monster in their game, and found that multiple DMs used a concept of ``baseline" monsters to gain a broad understanding of a monster when they first encounter it. 
The idea of the baseline monster was not to find a specific monster to compare another to, but to determine which parts of an individual monster's game statistics to focus on, and which parts to use prior thematic commonsense to fill in. 

In this context, we define \textit{thematic commonsense} as the DM's intuitive understanding of D\&D as a game with medieval fantasy themes, and how they might draw inspiration from other works of fantasy literature. For example, a DM might intuitively understand that a dragon is a kind of winged reptile with a fire breath based on their consumption of other fantasy works, reason that all dragons are capable of flight, and focus on a particular dragon's unique abilities rather than flight speed (D7). Although D\&D reference material does not include an explicit description of the dragon's fire breath, the DM might base their narration on depictions of fire breath from other authors.

We find this similar to the idea of a {\it genus-differentia} definition \cite{Parry_Hacker_1991}, in that DMs use their general background understanding of fantasy settings to define their personal {\it genus} and supplement prior knowledge by skimming monster reference books for {\it differentia}. This suggests that co-DM systems should focus on helping DMs extract these {\it differentiae}, and that they also require the same extensive background knowledge as the user. For the D\&D domain, we believe that LLMs such as GPT-3 \cite{GPT3} have included sufficient information on the game and the game books themselves in their training corpus so as to establish such a background knowledge. However, we are interested in methods for establishing this thematic commonsense knowledge for works not included in models' training data in future work.

\paragraph{Simple Language.}
Multiple DMs emphasized that they would like a co-DM system to present monster information in plain language, rather than the elaborate prose found in game reference manuals (D3-6). 
As a work of fantasy literature, D\&D publications (including reference manuals) often use heavy figurative language and obscure words. For example, the first paragraph of an owlbear's description reads:

\begin{quote}
    An owlbear’s screech echoes through dark valleys and benighted forests, piercing the quiet night to announce the death of its prey. Feathers cover the thick, shaggy coat of its bearlike body, and the limpid pupils of its great round eyes stare furiously from its owlish head \cite[pg. 147]{DnD_BR}.
\end{quote}

This style of description continues for seven additional paragraphs. On average, across all D\&D monsters published on D\&D Beyond, a monster's description and list of abilities contains 374 words (min: 0, max: 2,307). DMs often use multiple monsters together in the same encounter, compounding the amount of information they must hold in their mind.

Monster descriptions often include descriptions of the monster, its abilities, and lore. Some DMs' preferred method of referencing monster lore while running the game was to skim the full monster entry, and the complex and long prose often led to DMs feeling overwhelmed (D4, D5). Other DMs wanted a short and salient mechanical (i.e. focusing on monster's game abilities and actions) description, rather than a narrative (lore and history-focused) one (D3, D6).

Overall, the complexity of monster descriptions led DMs to forget parts of monsters' lore or abilities during gameplay (D5) or use overly broad simplifications that did not capture an individual monster's uniqueness (D6). While offline resources exist to help DMs run monsters (e.g. \citet{amman_monsters_know_2019}), they cannot account for the environment or generate a unique scenario for each encounter with the same monster. We believe that LLMs' capability to summarize and generate unique material is particularly applicable to these challenges.


\section{Implementation}
\label{sec:implementation}
\begin{figure*}[ht]
    \centering
    \begin{subfigure}[b]{0.325\textwidth}
        \centering
        \includegraphics[width=\textwidth]{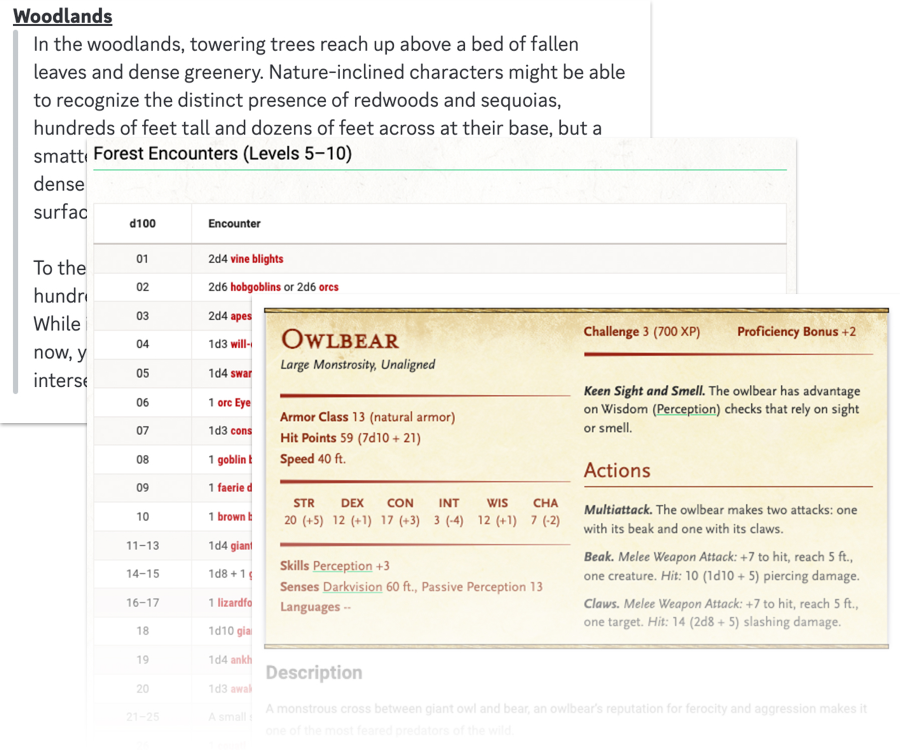}
        \caption{An example of the kinds of information DMs reference in an encounter. Digesting the myriad sources of information and using them to build a compelling narrative is a challenging task.}
        \label{fig:summary_info}
    \end{subfigure}
    \hfill
    \begin{subfigure}[b]{0.185\textwidth}
        \centering
        \includegraphics[width=\textwidth]{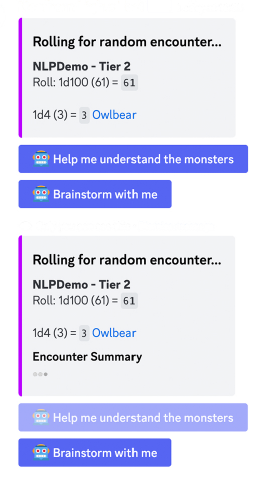}
        \caption{After rolling an encounter, the interface DMs used to interact with \calypso{}.}
        \label{fig:summary_thinking}
        \end{subfigure}
    \hfill
    \begin{subfigure}[b]{0.425\textwidth}
        \centering
        \includegraphics[width=\textwidth]{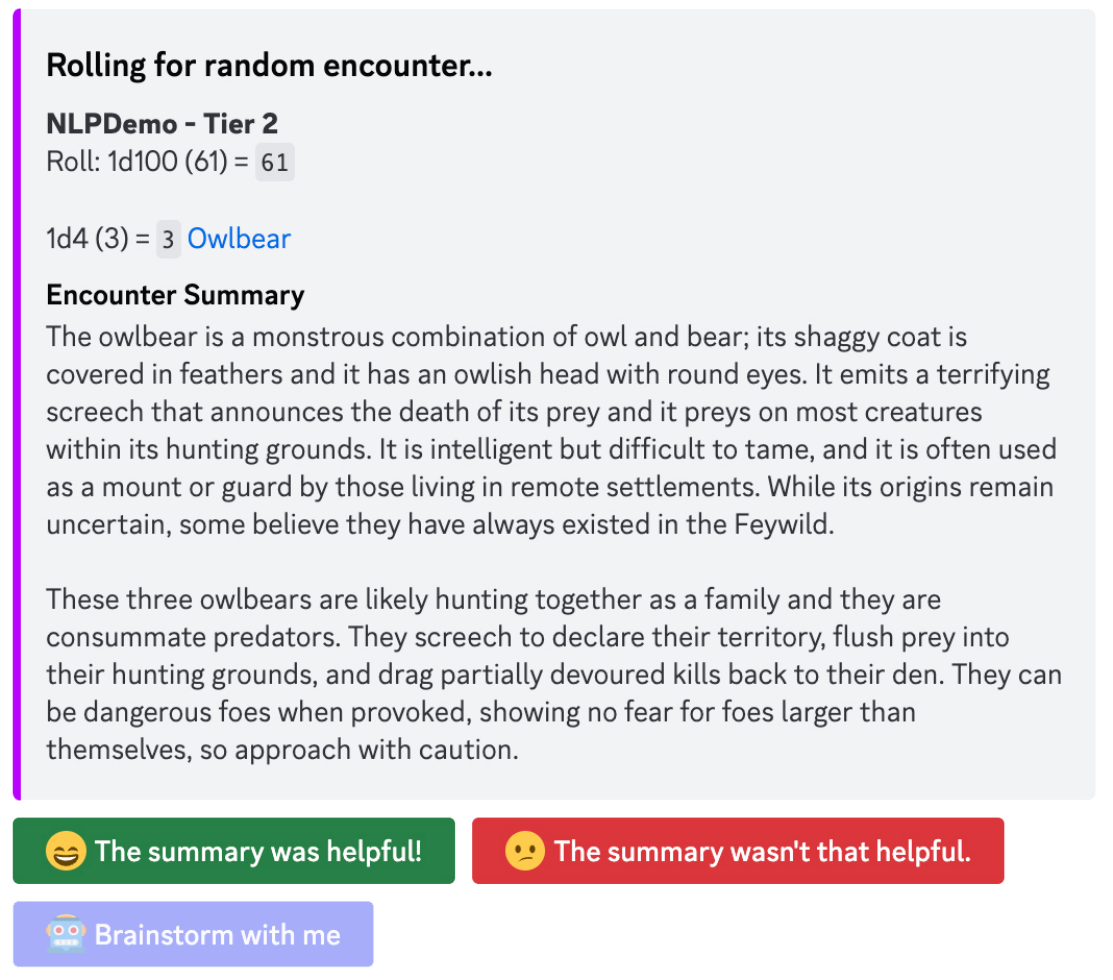}
        \caption{An example of \calypso{}'s summary of the encounter.}
        \label{fig:summary_gen}
    \end{subfigure}
    \caption{Using the {\em Encounter Understanding} interface to generate a distilled summary of an encounter.}
    \label{fig:summary}
\end{figure*}

In this section, we describe the three interfaces we developed to provide DMs with the sorts of support they desired. These interfaces were designed with "in the wild" deployment in mind:
\begin{enumerate}
    \item Encounter Understanding: a zero-shot method to generate a concise setup of an encounter, using GPT-3.
    \item Focused Brainstorming: a conversational method for DMs to ask additional questions about an encounter or refine an encounter summary, using ChatGPT.
    \item Open-Domain Chat Baseline: a conversational interface without the focus of an encounter, using ChatGPT.
\end{enumerate}

Our implementation differs from other efforts to develop AI-powered co-creative agents in two ways. First, compared to models where the AI acts as the writer, AI-generated content is not necessarily directly exposed to the audience. \calypso{} only presents ideas to a human DM, who has final say over what is presented to the players. Second, compared to co-writing assistants where the writer has plentiful time to iterate, the time between idea and presentation is very short. Since the DM uses \calypso{} in the midst of running a real game, \calypso{} should be frictionless to adopt and should not slow down the game.

\subsection{Encounter Understanding}
\label{sec:encounter-understanding}
The first interface we provided to DMs was a button to use a large language model to distill down game statistics and lore available in published monster stat blocks. To accomplish this, we prompted GPT-3 \cite{GPT3} (specifically, the text-davinci-003 model) with the text of the chosen encounter, the description of the setting the encounter was taking place in, and the game statistics and lore of each monster involved in the encounter. The full prompts are available in Appendix \ref{sec:prompts}. 

We began by presenting the LLM with the task to summarize monsters' abilities and lore and the environment. 
We collected feedback from DMs after generating the extracted information by allowing them to select a positive or negative feedback button, and optionally leave comments in an in-app modal. This interaction is illustrated in Figure \ref{fig:summary}.

\paragraph{Summarization.}
At first, we prompted GPT-3 to ``summarize the following D\&D setting and monsters for a DM's notes without mentioning game stats," then pasted verbatim the text description of the setting and monster information.
For decoding, we used a temperature of 0.9, top-p of 0.95, and frequency and presence penalties of 1. Based on feedback from DMs (discussed in Section \ref{sec:encounter-understanding-discussion}), we later changed to a more abstract ``understanding" task described below.

\paragraph{Abstractive Understanding.}
\label{sec:abstractive-understanding}
In the understanding task, we prompted GPT-3 with the more abstract task to help the DM ``understand" the encounter, along with explicit instructions to focus on the unique aspects of each creature, use information from mythology and common sense, and to mention how multiple creatures interact with each other. After these instructions, we included the same information as the {\em Summarization} task above. Finally, if a monster had no written description, we included instructions in place of the monster's description telling \calypso{} to provide the DM information from mythology and common sense.
For decoding, we used a temperature of 0.8, top-p of 0.95, and a frequency penalty of 0.5.

\subsection{Focused Brainstorming}
To handle cases where a single round of information extraction was not sufficient or a DM had additional focused questions or ideas they wanted assistance elaborating, we also provided an interface to open a private thread for focused brainstorming. Available at any time after an encounter was randomly chosen, we provided the same encounter information as in the {\em Encounter Understanding} interface as an initial prompt to ChatGPT (i.e., gpt-3.5-turbo) \cite{ChatGPT}. If the DM had used the {\em Encounter Understanding} interface to generate an information block, we also provided it as context (Figure \ref{fig:brainstorm}). The full prompts are available in Appendix \ref{sec:prompts}. For decoding, we used a temperature of 1, top-p of 0.95, and a frequency penalty of 0.3.

\subsection{Open-Domain Chat Baseline}
Finally, we made a baseline open-domain chat interface available to all players, without the focus of an encounter. As this interface was available at any time and open-ended, it helped provide a baseline for how DMs would use AI chatbots generally. To access the interface, users were able to run a bot command, which would start a new thread. We prompted ChatGPT to take on the persona of a fantasy creature knowledgeable about D\&D, and generated replies to every message sent in a thread opened in this manner. For decoding, we used a temperature of 1, top-p of 0.95, and a frequency penalty of 0.3. 
Unlike the private threads created by the {\em Focused Brainstorming} interface, open-domain conversation threads were public and allowed other users to join.

\section{Experimental Setup}
By deploying \calypso{} in the wild, we sought to learn how real DMs would adopt the new technology (if at all) and the emergent use cases that would arise.

We set up a special ``play-by-post living world'' game, which we describe below, and invited 71 players and DMs (referred to as P1-71) to participate by posting on D\&D recruitment forums. While preserving the core foundations of D\&D, our setup allowed us to conduct a large-scale study with a greater number of play sessions than studying individual games of D\&D.

In this section, we describe our methodology for setting up this large-scale D\&D game.

\subsection{D\&D Game Setup}
All gameplay occurred on our Discord server. We used Avrae, a D\&D Discord bot with over five million users, to facilitate gameplay. Avrae is commonly used to run D\&D games in this fashion, so the large-scale game was familiar to players and DMs \cite{FIREBALL}.
All participants were asked to review the server's research statement and to provide their informed consent before participating. Participants were compensated with free access to all published D\&D game materials (worth \$981.35). We explain the core differences between a traditional game of D\&D and our setup here:

\paragraph{Play-by-Post.}
While most commonly D\&D is played in person or using a virtual teleconference, a large number of players also play in a text-only mode known as ``play-by-post". In play-by-post games, rather than acting out characters using voices and body movements, players narrate their characters' actions and speech in a textual format. This text-based modality allowed us to monitor a large number of play sessions and allowed players to interface with language models without having to add an additional step to transcribe verbal play into text.

\paragraph{Living World.}
Our setup takes aspects from playing both prewritten modules and homebrew worlds. Traditionally, groups are comprised of 1 DM and 3-6 players playing in different worlds created by the DM, who play in regularly scheduled 3-4 hour play sessions (most commonly, once a week).
To allow for a larger scale study, in our setting, all 71 players exist in the same would, which we created. To emulate traditional play sessions, players form groups of 3-6 (on average) to partake in self-contained quests in the setting, always returning to a central hub after each quest. Within the hub, players are free to interact with each other, allowing room for storytelling and character development through roleplay without a DM. Outside the hub, we created a diverse set of environments that players could explore, each with a short description and image.

\subsection{Mode of Play}
In the \calypso{} study, the primary mode of play was a random encounter system: players explore the established world and the DM is given a set of monsters to use (randomly selected from a table of encounters). The DM must then create an encounter involving those monsters for their players to play in. 
By giving DMs access to the interfaces described above, we study how language models can be used with focused local knowledge to help a DM understand complex monster lore, interactions, and how to incorporate new encounters into an existing setting.
Figure \ref{fig:front} shows the flow of how an encounter is randomly chosen and how a DM might use \calypso{} to help build a compelling narrative.

\section{Results and Discussion}
In this section, we compile the ways that DMs adopted \calypso{}'s three interfaces into their games. We highlight the successes and failures of LLMs, and describe its emergent use cases.
We refer to the DMs who used \calypso{} as P1-P71 in this section.

\subsection{Encounter Understanding}
\label{sec:encounter-understanding-discussion}
\paragraph{Summarization.}
DMs interacted with the summarization model in 37 encounters, indicating that the summary helped them understand the monsters and setting in 13 encounters and did not help in 7 encounters.

We found that DMs appreciated the summary for distilling long information from stat blocks (P10) or highlighting an aspect of the monster to focus on (P8). However, when prompted with the task to summarize provided information, GPT-3 would focus too much on numeric game stats (despite the prompt stating not to) and summarize the environment and monsters separately, often focusing too much on the environment (P3). For example, for an encounter involving 12 Blink Dogs, the model's output simply paraphrased the information written in the setting and stat block rather than suggesting more interesting interactions:

\begin{quote}
    {\bf \calypso{}}: A warm and calm autumnal forest with a river running through it, the players can travel through this area towards the distant mountain range. The creatures they may encounter here are blink dogs; they have an armor class of 13, 22 Hit Points, a speed of 40 ft., as well as skills in Perception and Stealth. They also possess limited intelligence and understand Sylvan though cannot speak it.
\end{quote}

Although the information presented by \calypso{} is factually correct, DMs argued that simply rephrasing the information in the monster's reference material ``doesn't really help in running an encounter" (P8) and generally wanted ``more information about [the monsters] and why they might interact with [the players]" (P11). DMs wanted \calypso{} to generate details that were entirely omitted from the reference material, using the written reference as only a basis to gain an understanding of the monster.

Another common case where \calypso{} did not present a helpful summary was when a monster did not have a published prose description, only numeric game stats. Most commonly, such creatures were an animal found in real life (e.g. a brown bear; P3) or relied entirely on a DM's thematic commonsense to run (e.g. a snow golem; P3). In these cases, the model would only use minimal external knowledge (e.g. noting that giant boars usually had ``thick, protective skin") rather than using thematic commonsense to make up a more interesting narrative (discussed in Section \ref{sec:thematic-commonsense}). We found that in order for the LLM to use external knowledge about monsters rather than only what was written in game stats, we had to prompt it with explicit instructions to do so. With these findings in mind, we remodeled the summarization task to a more abstract ``understanding" task, in which we provided the model explicit instructions to use thematic commonsense.

\begin{figure}
    \centering
    \includegraphics[width=\columnwidth]{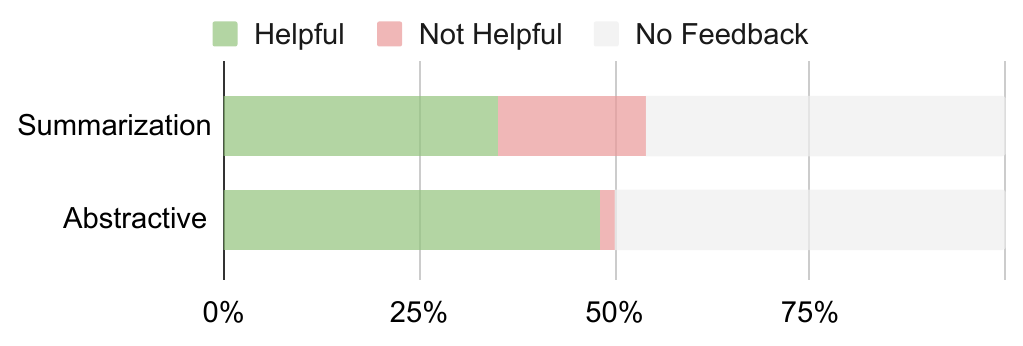}
    \caption{DMs found the {\em Abstractive Understanding} method of distilling monster information more consistently helpful than the {\em Summarization} method.}
    \label{fig:summary_feedback}
\end{figure}

\paragraph{Abstractive Understanding.}
\label{sec:understanding-discussion}
Overall, DMs much preferred this more abstract task: over interactions in 114 encounters, DMs indicated that the summary helped them understand the monsters and setting in 55 encounters and did not help in 2 encounters. Figure \ref{fig:summary_feedback} shows the breakdown of feedback for each interface.

Primarily, DMs using this model were satisfied with its ability to describe monsters based on real-life animals (P8), creature behavior (P3), interactions with players (P3), how to roleplay creatures (P12), and general performance (P3, P8, P11-13). The most common undesired behavior DMs noticed was the model hallucinating, or making up, facts about the monsters, such as giving giant spiders telepathy (P15), an insect monster a poison ability (P8), or a frost salamander wings (P14). Given the same example encounter as above, the model's output includes much more useful information about appearance and monster behavior:

\begin{quote}
    {\bf \calypso{}}: The blink dogs are canine creatures, about the size of a large dog. Their fur is a mottled grey and white colour, and their eyes flash yellow when they use their supernatural ability to blink in and out of existence. They are surprisingly fast despite their size, and can easily keep up with most horses. In terms of behavior, the blink dogs are curious but shy creatures who prefer to avoid contact with other creatures unless absolutely necessary. If threatened, they will use their blinking power to confuse and disorientate their opponents before attacking with a pack mentality. The blink dogs of this forest seem particularly adept at working together as a unit and can prove a formidable foe if provoked.
\end{quote}

As the {\em Summarization} and {\em Abstractive Understanding} tasks differ only in prompting methods, we demonstrate how sensitive prompting approaches can be. It is possible that GPT-3 is overfit on the word ``summarize", and even subtle changes in prompt design encourage original and domain-specific generation.
With these alterations, we found that LLMs are capable of synthesizing creative ideas about D\&D monsters using thematic commonsense contained in its training priors.

We also found that if we repeated a certain phrase multiple times in the prompt, e.g. ``you may use information from mythology and common sense" (which could occur if an encounter involved multiple monsters with no published lore), the model would often include the phrase ``mythology and common sense" verbatim in its output. This is likely a case of degenerative output \cite{Holtzman2020Degeneration}. To prevent this, each time a set phrase would have been templated in, we instead randomly chose 2-4 words from the set \{"folklore", ``common\ sense", ``mythology", ``culture"\} and inserted them in a random order (e.g. ``use information from common sense, mythology, and folklore").

\paragraph{Effect of Hallucinations.}
We find that not all hallucinations are undesired. In many cases, the model suggests monster behaviors or appearances that are not explicitly written out in monster descriptions, such as the appearance of the blink dogs' fur in the example above. More drastic deviations, such as the model suggesting giving a creature wings, were however undesired.

DMs often take creative liberty to synthesize sensical information that isn't included in the source material. As shown above, they expect their tools to do the same when necessary -- while the \textit{Summarization} interface was more conservative in ensuring it did not hallucinate any details, the \textit{Abstractive Understanding} interface was more well-received even with minor hallucinations. Since the DM acts as a curator of the model's output, the DM can choose which of the generations to accept.


\subsection{Focused Brainstorming}

\begin{figure}
    \centering
    \includegraphics[width=\columnwidth]{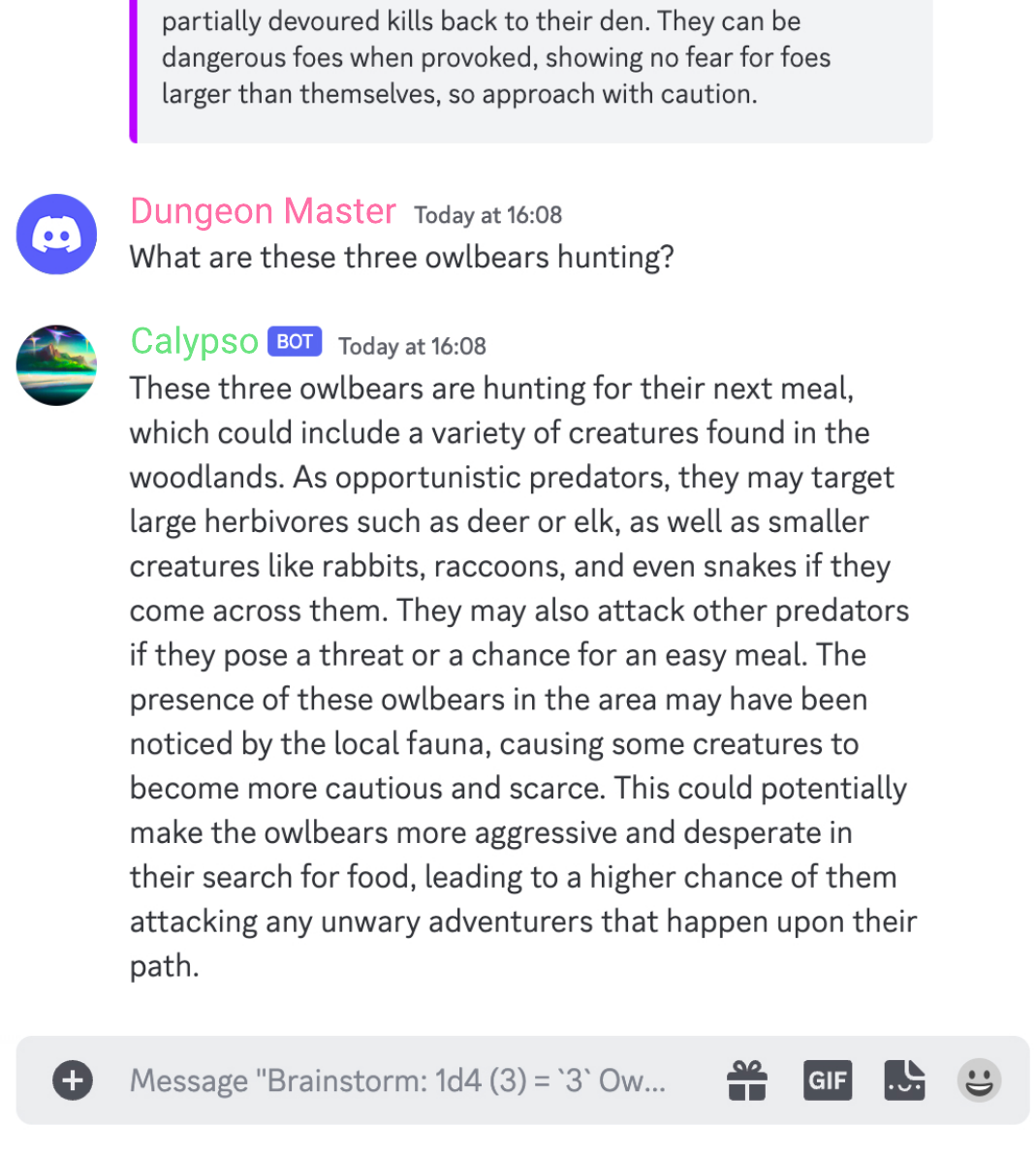}
    \caption{Using the {\em Focused Brainstorming} interface to ask specific questions about an encounter. \calypso{} suggests reasons why the players might encounter the monsters and how they might act.}
    \label{fig:brainstorm}
\end{figure}

{
\renewcommand{\arraystretch}{1.3}
\begin{table*}[ht]
\begin{tabular}{l p{6.0cm} p{7.4cm}} 
 \hline
 \textbf{Use Case} & \textbf{Description} & \textbf{Example} \\
 \hline
 General Descriptions & Asking the model to generate a high-level description of a scene and encounter. & {\em ``Describe this encounter from the player's perspective."} (P8) \\ 
 Specific Descriptions & Asking specific questions about parts of the encounter, often in response to player actions. & {\em ``Describe drinking games that the satyrs are taking part in that are so dangerous someone could get hurt doing them."} (P17) \\ 
 Strategy & Using the model to understand monster motives and get suggestions for their tactics. & {\em ``Why would a Displacer Beast Kitten leave the safety of its den if it believes an intruder is nearby?"} (P12) \\
 Making Decisions & Using the model to decide how the DM should run a given encounter. & {\em ``Should a diplomatic solution be possible for this encounter?"} (P14) \\
 List of Ideas & Generating a list of multiple ideas to build off of individually. & {\em ``give me encounter ideas"} (P10) \newline {\em ``...make up more [magic items] to make this encounter more interesting."} (P19) \\
 \hline
\end{tabular}
\caption{A list of common ways DMs used the {\em Focused Brainstorming} interface.}
\label{tab:brainstorm}
\end{table*}
}

In total, DMs used the focused brainstorming model in 71 encounters, comprising a total of 162 rounds of conversation.
DMs used the brainstorming model in a number of diverse way, which we qualitatively coded and tabulate in Table \ref{tab:brainstorm}. Here, we discuss these use cases and some failure cases.

\paragraph{General and Specific Descriptions.}
The most common way DMs used the interface was to ask it for a high level description of a given encounter and specific descriptions of points in the encounter. Since our prompt included information on the setting and involved monsters, the model was able to reference the information in its description. Additionally, the conversational nature of the language model added to its context, so DMs could reference earlier ideas without having to repeat them. This allowed DMs to ask \calypso{} to simply ``describe this scene" or ``describe X" without having to specify additional details (P3, P8-10, P12, P16-20).

After presenting an idea to their players and seeing what part of the encounter players interacted with, the DM was also able to ask follow-up questions to describe in detail specific elements the players interacted with. For example, when running an encounter involving a ship's figurehead that was washed ashore, P3 first asked for a description of the figurehead. Then, when the players investigated it further, the DM followed up by asking for ``a description about its construction, specifically how it was carved, and perhaps what D\&D race crafted it." This allowed DMs to elaborate on specific parts of an encounter when it became relevant, rather than presenting a large amount of information up front.

However, DMs found that the model struggled sometimes to describe combat, and suggested that including more information about the combat state (similar to \citet{FIREBALL}) or map placement information could help generate more specific descriptions (P3, P9). Some DMs used these descriptions verbatim (P3, P8, P17), while others picked out particularly vivid phrases to use in a description of their own (P3, P8, P10, P12, P20). Others disagreed with the model's description and wrote their own instead (P13, P16, P18, P19).

\paragraph{Strategy.}
Another common use case for DMs was to ask the model for monsters' ``motives, tactics, and who they might prioritize [in a fight]" (P8-9, P12-13, P19, P23). As discussed in section \ref{sec:strategic-copilot} ({\em Strategic Copilot}), coming up with and sticking to strategies for each monster can be overwhelming, and often DMs use simplifications to manage their mental load. This use case allowed DMs to create more engaging fights with clearer paths to resolutions by describing a creature's motive and specific tactics the creature would use. For example, when a DM asked how a pack of ten wolves might approach a camping party, the model suggested to have the wolves ``circle around the camp, hiding behind trees and bushes [...] and wait until a member of the party is alone and vulnerable before striking, hoping to separate and weaken the group" (P8). Similar to the interactions with descriptions, these DMs did not always use the strategy presented by the model; sometimes they picked and chose interesting suggestions, while other times they chose a different approach.

\paragraph{Making Decisions.}
Some DMs used the model to get an opinion on two options they had already written or thought of (P3, P8-9, P12-14, P18, P23). For example, when players encountered a ravine whose bottom was enshrouded in mist, one DM asked whether the mist should hide a very long or short drop. The model would sometimes simply give feedback on both of the options without choosing one ({\em ``Both options have their merits depending on the tone and style of your game..."}; P3) and sometimes give a more straightforward answer (``...would that revenant have a vengeance towards the party member?" / {\em ``Yes, absolutely..."}; P12). DMs did not ask the model to come to a conclusive decision, suggesting that the model providing its ``opinion" helped inspire the DM, without relying on it to run the encounter.

\paragraph{List of Ideas.}
In this use case, the DM simply asks the model for a list of ideas; for example, a list of magic items sea-dwelling humanoids might have (P10). We believe that the reasoning for this use case is the same reason that makes random tables (as discussed in Section \ref{sec:dnd}) a popular method of inspiration -- however, compared to prewritten random tables, LLMs have the powerful capability of generating unique ``random table" entries customized for specific contexts.

\paragraph{Failure Cases.}
The most common failure case was when DMs tried to invoke other tools (such as a dice rolling or spell search bot) available in the brainstorming chat.
As the model responded to every message in the thread, it would also respond to the other tool's invocation and reply with a generic error message or try to infer the other tool's output (e.g. ``!check stealth" / \textit{``Abominable Yeti stealth check: 18"}, hallucinating a result while ignoring the output of an actual dice roller). In some cases, the DM attempted to upload an image, which the model was unable to view. Finally, as discussed in Section \ref{sec:understanding-discussion}, the model sometimes hallucinated facts about creatures and rules. We believe multimodality (allowing the model to view images) and allowing the model to use tools (e.g. to retrieve rules text, spell descriptions, or search monsters) to be an interesting direction to explore in future work.

We also find that certain artifacts of the model's training process influences its output. For example, the model would sometimes refuse to suggest (fantasy) races, likely due to efforts to reduce the potential for real-world racial bias. In another case, the model insists that it is incapable of playing D\&D, likely due to efforts to prevent the model from making claims of abilities it does not possess. Although generally infrequent, these artifacts suggest that domain-specific fine-tuning may improve models' performance.

\subsection{Open-Domain Chat Baseline}
Participants chatted with \calypso{} in 51 unique threads, comprising a total of 2,295 rounds of conversation. Compared to conversations with the AI in the {\em Focused Brainstorming} interface, conversations lasted much longer (averaging 45.0 rounds per interaction vs. the brainstorming interface's 2.3).
Without the time pressure of an active game that the DM is responsible for, participants spent more time playing with the model and refining its responses to generate high-level quest ideas (P3, P8, P12, P16), character and location names (P3, P9, P19, P22), role-play specific characters from other games (P3, P9, P12, P16), and write fanfiction about events happening between their characters in the game (P3, P8, P9, P16, P21), among other non-D\&D uses.

However, during a game of D\&D, DMs did not have the time luxury to iterate on responses for hours. Without \calypso{}'s management of the game, DMs would have to spend many turns of conversation copying and pasting information to provide it to the LLM, taking attention away from the game and making the baseline implementation unsuitable for real-world adoption.

We believe this highlights the difference between synchronous and asynchronous systems and the importance of removing friction from AI-augmented user interfaces as discussed in Section \ref{sec:implementation} -- while the human user may have the capability to supply a LLM with additional information, the time and computational burden should be on the synchronous system rather than the user.


\section{Conclusions}
In this paper, we present \calypso{}, a system of three LLM-powered interfaces that DMs could use to assist them in preparing and running focused monster encounters in an established setting, and a large-scale study of how 71 D\&D players incorporated \calypso{} into their gameplay. Through interviews with DMs, we established common themes and desires for AI-augmented DM tools, and used these motivations and iterative design to guide our development. In conclusion, we found that:
\begin{enumerate}
    \item LLMs are capable brainstorming partners. DMs used \calypso{} to generate both low-fidelity ideas that they could grow using their own creative expression, and guided it to generate high-fidelity descriptions they could present to other players with only minor edits.
    \item LLMs present thematic commonsense when prompted to. Having been trained on a large corpus containing D\&D texts and discussions, works of fantasy literature, and descriptions of real-world creatures, \calypso{} was able to fill in gaps in the D\&D literature by probing into thematically relevant common sense knowledge. However, we found that to access this trove of information, the LLM had to be explicitly prompted to do so.
    \item LLMs assist, rather than replace, human DMs. \calypso{} was designed to aid a human DM while maintaining their creative agency. We find that human DMs use AI co-DMs to understand complex rules text, brainstorm interactions between non-player characters or monsters, and present DMs with suggestions that the DM can weave into a story to present to players without taking away from the pace of the game. Human creativity is an integral part of storytelling games like D\&D, and it is important for future AI tools to always maintain the human's creative agency.
\end{enumerate}

\appendix





\section{LLM Prompts}
\label{sec:prompts}
In this section, we provide the prompts used in the \calypso{} system. Generally, we make use of Markdown-style headers to divide sections of the prompt. For chat-based models, we annotate each message with the corresponding role (system, assistant, or user, as exposed in the ChatGPT API).

\subsection{Encounter Understanding}
\paragraph{Summarization}
\leavevmode
\begin{lstlisting}
Summarize the following D&D setting and monsters for a Dungeon Master's notes without mentioning game stats.

Setting
=======
<Setting description inserted here.>

Creatures
=========
<Name>
------
<Statistics and lore inserted here. If the encounter involves multiple creatures, repeat for each creature.>

Summary
=======
\end{lstlisting}

\paragraph{Abstractive Understanding}
\leavevmode
\begin{lstlisting}
Your name is Calypso, and your job is to help the Dungeon Master with an encounter. 
Your task is to help the DM understand the setting and creatures as a group, focusing mainly on appearance and how they act.
Especially focus on what makes each creature stand out.
Avoid mentioning game stats.
You may use information from common sense, mythology, and culture.
If there are multiple creatures, conclude by mentioning how they interact.

Encounter: <Encounter inserted here.>
\end{lstlisting}
The rest of the prompt follows as in the Summarization prompt above, beginning with the setting. If a monster did not have published lore, we inserted the string ``Calypso, please provide the DM with information about the (monster name) using information from (folklore, common sense, mythology, and culture)" (see section \ref{sec:understanding-discussion}) in place of lore.

\subsection{Focused Brainstorming}
\begin{lstlisting}
SYSTEM: You are a creative D&D player and DM named Calypso.
Avoid mentioning game stats. You may use information from common sense, mythology, and culture.

USER: I'm running this D&D encounter: <Encounter inserted here.>

<Setting and creatures inserted here, in the same format as Abstractive Understanding.>

Your job is to help brainstorm some ideas for the encounter.
\end{lstlisting}
If the DM used the Encounter Understanding interface before starting a brainstorming thread, we add an additional message to the prompt:
\begin{lstlisting}
USER: Here's what I have so far:
<Summary generated by Encounter Understanding inserted here.>
\end{lstlisting}
This allows the DM to reference ideas proposed by \calypso{} in its summary without having to repeat the entire message, aiding continuity.



\section*{Acknowledgments}
Thank you to the Northern Lights Province Discord server for playing with us and being so enthusiastic about AI and D\&D! Thank you to the NLP server staff - friends and players who helped us write rules, settings, game mechanics, and manage so many players: Ryan Crowley, Nicki Dulmage-Bekker, @ephesia, @lyra.kat, and Joseph Keen. Finally, thank you to D\&D Beyond for providing us with access to monster information and game materials.

This material is based upon work supported by the National Science Foundation under Grant \#2030859 to the Computing Research Association for the CIFellows Project.

\bibliography{aaai23, references, zotero}

\begin{thebibliography}{40}
\providecommand{\natexlab}[1]{#1}

\bibitem[{Acharya, Mateas, and Wardrip-Fruin(2021)}]{acharya_interviews_2021}
Acharya, D.; Mateas, M.; and Wardrip-Fruin, N. 2021.
\newblock Interviews {Towards} {Designing} {Support} {Tools} for {TTRPG}
  {Game} {Masters}.
\newblock In Mitchell, A.; and Vosmeer, M., eds., \emph{Interactive
  {Storytelling}}, Lecture {Notes} in {Computer} {Science}, 283--287. Cham:
  Springer International Publishing.
\newblock ISBN 978-3-030-92300-6.

\bibitem[{Akoury et~al.(2020)Akoury, Wang, Whiting, Hood, Peng, and
  Iyyer}]{akoury_storium_2020}
Akoury, N.; Wang, S.; Whiting, J.; Hood, S.; Peng, N.; and Iyyer, M. 2020.
\newblock {STORIUM}: {A} {Dataset} and {Evaluation} {Platform} for
  {Machine}-in-the-{Loop} {Story} {Generation}.
\newblock In \emph{Proceedings of the 2020 {Conference} on {Empirical}
  {Methods} in {Natural} {Language} {Processing} ({EMNLP})}, 6470--6484.
  Online: Association for Computational Linguistics.

\bibitem[{Amman(2019)}]{amman_monsters_know_2019}
Amman, K. 2019.
\newblock \emph{The Monsters Know What They're Doing}.
\newblock New York, NY: Gallery Books.
\newblock ISBN 9781982122669.

\bibitem[{Ammanabrolu et~al.(2020)Ammanabrolu, Cheung, Tu, Broniec, and
  Riedl}]{ammanabrolu_bringing_2020}
Ammanabrolu, P.; Cheung, W.; Tu, D.; Broniec, W.; and Riedl, M. 2020.
\newblock Bringing {Stories} {Alive}: {Generating} {Interactive} {Fiction}
  {Worlds}.
\newblock \emph{Proceedings of the AAAI Conference on Artificial Intelligence
  and Interactive Digital Entertainment}, 16(1): 3--9.

\bibitem[{Bergström(2011)}]{bergstrom_framing_2011}
Bergström, K. 2011.
\newblock Framing {Storytelling} with {Games}.
\newblock In Si, M.; Thue, D.; André, E.; Lester, J.~C.; Tanenbaum, T.~J.; and
  Zammitto, V., eds., \emph{Interactive {Storytelling}}, Lecture {Notes} in
  {Computer} {Science}, 170--181. Berlin, Heidelberg: Springer.
\newblock ISBN 978-3-642-25289-1.

\bibitem[{Brown et~al.(2020)Brown, Mann, Ryder, Subbiah, Kaplan, Dhariwal,
  Neelakantan, Shyam, Sastry, Askell, Agarwal, Herbert-Voss, Krueger, Henighan,
  Child, Ramesh, Ziegler, Wu, Winter, Hesse, Chen, Sigler, Litwin, Gray, Chess,
  Clark, Berner, McCandlish, Radford, Sutskever, and Amodei}]{GPT3}
Brown, T.; Mann, B.; Ryder, N.; Subbiah, M.; Kaplan, J.~D.; Dhariwal, P.;
  Neelakantan, A.; Shyam, P.; Sastry, G.; Askell, A.; Agarwal, S.;
  Herbert-Voss, A.; Krueger, G.; Henighan, T.; Child, R.; Ramesh, A.; Ziegler,
  D.; Wu, J.; Winter, C.; Hesse, C.; Chen, M.; Sigler, E.; Litwin, M.; Gray,
  S.; Chess, B.; Clark, J.; Berner, C.; McCandlish, S.; Radford, A.; Sutskever,
  I.; and Amodei, D. 2020.
\newblock Language Models are Few-Shot Learners.
\newblock In Larochelle, H.; Ranzato, M.; Hadsell, R.; Balcan, M.; and Lin, H.,
  eds., \emph{Advances in Neural Information Processing Systems}, volume~33,
  1877--1901. Curran Associates, Inc.

\bibitem[{Calderwood et~al.(2020)Calderwood, Qiu, Gero, and
  Chilton}]{calderwood_how_2020}
Calderwood, A.; Qiu, V.; Gero, K.~I.; and Chilton, L.~B. 2020.
\newblock How {Novelists} {Use} {Generative} {Language} {Models}: {An}
  {Exploratory} {User} {Study}.
\newblock In \emph{International {Conference} on {Intelligent} {User}
  {Interfaces} ({IUI}) {Workshops}}. Cagliari, Italy: ACM.

\bibitem[{Callison-Burch et~al.(2022)Callison-Burch, Singh~Tomar, Martin,
  Ippolito, Bailis, and Reitter}]{Callison-Burch2022}
Callison-Burch, C.; Singh~Tomar, G.; Martin, L.~J.; Ippolito, D.; Bailis, S.;
  and Reitter, D. 2022.
\newblock {Dungeons and Dragons as a Dialogue Challenge for Artificial
  Intelligence}.
\newblock In \emph{Conference on Empirical Methods in Natural Language
  Processing (EMNLP)}, 9379–9393. Abu Dhabi, United Arab Emirates:
  Association for Computational Linguistics.

\bibitem[{Chung et~al.(2022)Chung, Kim, Yoo, Lee, Adar, and
  Chang}]{chung_talebrush_2022}
Chung, J. J.~Y.; Kim, W.; Yoo, K.~M.; Lee, H.; Adar, E.; and Chang, M. 2022.
\newblock {TaleBrush}: {Sketching} {Stories} with {Generative} {Pretrained}
  {Language} {Models}.
\newblock In \emph{{CHI} {Conference} on {Human} {Factors} in {Computing}
  {Systems}}, 1--19. New Orleans LA USA: ACM.
\newblock ISBN 978-1-4503-9157-3.

\bibitem[{Coenen et~al.(2021)Coenen, Davis, Ippolito, Reif, and
  Yuan}]{coenen_wordcraft_2021}
Coenen, A.; Davis, L.; Ippolito, D.; Reif, E.; and Yuan, A. 2021.
\newblock Wordcraft: a {Human}-{AI} {Collaborative} {Editor} for {Story}
  {Writing}.
\newblock In \emph{First {Workshop} on {Bridging} {Human}-{Computer}
  {Interaction} and {Natural} {Language} {Processing} at {EACL} 2021}.
  Association for Computational Linguistics.

\bibitem[{Crawford, Mearls, and Perkins(2018)}]{DnD_BR}
Crawford, J.; Mearls, M.; and Perkins, C. 2018.
\newblock \emph{D\&D Basic Rules}.
\newblock Renton, WA: Wizards of the Coast.

\bibitem[{Crawford, Perkins, and Wyatt(2014)}]{DnD_DMG}
Crawford, J.; Perkins, C.; and Wyatt, J. 2014.
\newblock \emph{Dungeon Master's Guide}.
\newblock Renton, WA: Wizards of the Coast.

\bibitem[{{D\&D Beyond}(2017)}]{DnDBeyond}
{D\&D Beyond}. 2017.
\newblock D\&D Beyond.
\newblock \url{https://www.dndbeyond.com/}.

\bibitem[{dScryb(2020)}]{dScryb}
dScryb. 2020.
\newblock dScryb.
\newblock \url{https://dscryb.com/}.

\bibitem[{{Foundry Gaming, LLC}(2019)}]{FoundryVTT}
{Foundry Gaming, LLC}. 2019.
\newblock Foundry Virtual Tabletop.
\newblock \url{https://foundryvtt.com/}.

\bibitem[{Gygax and Arneson(1974)}]{Gygax1974}
Gygax, G.; and Arneson, D. 1974.
\newblock \emph{{Dungeons \& Dragons}}.

\bibitem[{Holtzman et~al.(2020)Holtzman, Buys, Du, Forbes, and
  Choi}]{Holtzman2020Degeneration}
Holtzman, A.; Buys, J.; Du, L.; Forbes, M.; and Choi, Y. 2020.
\newblock The Curious Case of Neural Text Degeneration.
\newblock In \emph{International Conference on Learning Representations}.

\bibitem[{Ippolito et~al.(2022)Ippolito, Yuan, Coenen, and
  Burnam}]{ippolito_creative_2022}
Ippolito, D.; Yuan, A.; Coenen, A.; and Burnam, S. 2022.
\newblock Creative {Writing} with an {AI}-{Powered} {Writing} {Assistant}:
  {Perspectives} from {Professional} {Writers}.
\newblock ArXiv:2211.05030 [cs].

\bibitem[{Kelly, Mateas, and Wardrip-Fruin(2023)}]{kelly_towards_2023}
Kelly, J.; Mateas, M.; and Wardrip-Fruin, N. 2023.
\newblock Towards {Computational} {Support} with {Language} {Models} for
  {TTRPG} {Game} {Masters}.
\newblock In \emph{Proceedings of the 18th {International} {Conference} on the
  {Foundations} of {Digital} {Games}}, {FDG} '23, 1--4. New York, NY, USA:
  Association for Computing Machinery.
\newblock ISBN 978-1-4503-9855-8.

\bibitem[{Kreminski et~al.(2022)Kreminski, Dickinson, Wardrip-Fruin, and
  Mateas}]{kreminski_loose_2022}
Kreminski, M.; Dickinson, M.; Wardrip-Fruin, N.; and Mateas, M. 2022.
\newblock Loose {Ends}: {A} {Mixed}-{Initiative} {Creative} {Interface} for
  {Playful} {Storytelling}.
\newblock \emph{Proceedings of the AAAI Conference on Artificial Intelligence
  and Interactive Digital Entertainment}, 18(1): 120--128.
\newblock Number: 1.

\bibitem[{Lin, Agarwal, and Riedl(2022)}]{lin_creative_2022}
Lin, Z.; Agarwal, R.; and Riedl, M. 2022.
\newblock Creative {Wand}: {A} {System} to {Study} {Effects} of
  {Communications} in {Co}-{Creative} {Settings}.
\newblock \emph{AAAI Conference on Artificial Intelligence and Interactive
  Digital Entertainment (AIIDE)}, 18(1): 45--52.

\bibitem[{Louis and Sutton(2018)}]{Louis2018}
Louis, A.; and Sutton, C. 2018.
\newblock Deep {Dungeons} and {Dragons}: {Learning} {Character}-{Action}
  {Interactions} from {Role}-{Playing} {Game} {Transcripts}.
\newblock In \emph{Conference of the {North} {American} {Chapter} of the
  {Association} for {Computational} {Linguistics}: {Human} {Language}
  {Technologies} ({NAACL}-{HLT})}, volume Volume 2 (Short Papers), 708--713.
  New Orleans, Louisiana: Association for Computational Linguistics.

\bibitem[{Martin, Sood, and Riedl(2018)}]{Martin2018}
Martin, L.~J.; Sood, S.; and Riedl, M.~O. 2018.
\newblock {Dungeons and DQNs: Toward Reinforcement Learning Agents that Play
  Tabletop Roleplaying Games}.
\newblock In Wu, H.-Y.; Si, M.; and Jhala, A., eds., \emph{Joint Workshop on
  Intelligent Narrative Technologies and Workshop on Intelligent Cinematography
  and Editing (INT-WICED)}. Edmonton, AB, Canada: http://ceur-ws.org.

\bibitem[{Newman and Liu(2022)}]{newman_generating_2022}
Newman, P.; and Liu, Y. 2022.
\newblock Generating {Descriptive} and {Rules}-{Adhering} {Spells} for
  {Dungeons} \& {Dragons} {Fifth} {Edition}.
\newblock In \emph{Proceedings of the 9th {Workshop} on {Games} and {Natural}
  {Language} {Processing} within the 13th {Language} {Resources} and
  {Evaluation} {Conference}}, 54--60. Marseille, France: European Language
  Resources Association.

\bibitem[{OpenAI(2022)}]{ChatGPT}
OpenAI. 2022.
\newblock Introducing ChatGPT.
\newblock \url{https://openai.com/blog/chatgpt}.

\bibitem[{Parry and Hacker(1991)}]{Parry_Hacker_1991}
Parry, W.~T.; and Hacker, E.~A. 1991.
\newblock \emph{Aristotelian logic}.
\newblock Albany, NY: State University of New York Press.
\newblock ISBN 9780791406892.

\bibitem[{Perez, Eisemann, and Bidarra(2021)}]{perez_synset-based_2021}
Perez, M. R.~B.; Eisemann, E.; and Bidarra, R. 2021.
\newblock A {Synset}-{Based} {Recommender} {Method} for {Mixed}-{Initiative}
  {Narrative} {World} {Creation}.
\newblock In Mitchell, A.; and Vosmeer, M., eds., \emph{Interactive
  {Storytelling}}, Lecture {Notes} in {Computer} {Science}, 13--28. Cham:
  Springer International Publishing.
\newblock ISBN 978-3-030-92300-6.

\bibitem[{Perkins et~al.(2014)Perkins, Crawford, Sims, Thompson, Lee, Mearls,
  Schwalb, Sernett, Townshend, and Wyatt}]{DnD_MM}
Perkins, C.; Crawford, J.; Sims, C.; Thompson, R.; Lee, P.; Mearls, M.;
  Schwalb, R.~J.; Sernett, M.; Townshend, S.; and Wyatt, J. 2014.
\newblock \emph{Monster Manual}.
\newblock Renton, WA: Wizards of the Coast.

\bibitem[{Rameshkumar and Bailey(2020)}]{Rameshkumar2020}
Rameshkumar, R.; and Bailey, P. 2020.
\newblock Storytelling with {Dialogue}: {A} {Critical} {Role} {Dungeons} and
  {Dragons} {Dataset}.
\newblock In \emph{Annual {Meeting} of the {Association} for {Computational}
  {Linguistics} ({ACL})}, 5121--5134. Online: Association for Computational
  Linguistics.

\bibitem[{Roemmele and Gordon(2015)}]{schoenau-fog_creative_2015}
Roemmele, M.; and Gordon, A.~S. 2015.
\newblock Creative {Help}: {A} {Story} {Writing} {Assistant}.
\newblock In Schoenau-Fog, H.; Bruni, L.~E.; Louchart, S.; and Baceviciute, S.,
  eds., \emph{Interactive {Storytelling}}, volume 9445, 81--92. Cham: Springer
  International Publishing.
\newblock ISBN 978-3-319-27035-7 978-3-319-27036-4.
\newblock Series Title: Lecture Notes in Computer Science.

\bibitem[{Roven(2014)}]{TabletopAudio}
Roven, T. 2014.
\newblock Tabletop Audio.
\newblock \url{https://tabletopaudio.com/}.

\bibitem[{Samuel, Mateas, and Wardrip-Fruin(2016)}]{nack_design_2016}
Samuel, B.; Mateas, M.; and Wardrip-Fruin, N. 2016.
\newblock The {Design} of {Writing} {Buddy}: {A} {Mixed}-{Initiative}
  {Approach} {Towards} {Computational} {Story} {Collaboration}.
\newblock In Nack, F.; and Gordon, A.~S., eds., \emph{Interactive
  {Storytelling}}, volume 10045, 388--396. Cham: Springer International
  Publishing.
\newblock ISBN 978-3-319-48278-1 978-3-319-48279-8.
\newblock Series Title: Lecture Notes in Computer Science.

\bibitem[{Santiago et~al.(2023)Santiago, Parayno, Deja, and
  Samson}]{santiago_iii_rolling_2023}
Santiago, J.~M., III; Parayno, R.~L.; Deja, J.~A.; and Samson, B. P.~V. 2023.
\newblock Rolling the {Dice}: {Imagining} {Generative} {AI} as a {Dungeons} \&
  {Dragons} {Storytelling} {Companion}.
\newblock ArXiv:2304.01860 [cs].

\bibitem[{van Velsen, Williams, and
  Verhulsdonck(2009)}]{van_velsen_table-top_2009}
van Velsen, M.; Williams, J.; and Verhulsdonck, G. 2009.
\newblock Table-{Top} {Gaming} {Narratology} for {Digital} {Interactive}
  {Storytelling}.
\newblock In Iurgel, I.~A.; Zagalo, N.; and Petta, P., eds., \emph{Interactive
  {Storytelling}}, Lecture {Notes} in {Computer} {Science}, 109--120. Berlin,
  Heidelberg: Springer.
\newblock ISBN 978-3-642-10643-9.

\bibitem[{Vaswani et~al.(2017)Vaswani, Shazeer, Parmar, Uszkoreit, Jones,
  Gomez, Kaiser, and Polosukhin}]{transformer}
Vaswani, A.; Shazeer, N.; Parmar, N.; Uszkoreit, J.; Jones, L.; Gomez, A.~N.;
  Kaiser, L.; and Polosukhin, I. 2017.
\newblock Attention Is All You Need.
\newblock arXiv:1706.03762.

\bibitem[{Yang et~al.(2022)Yang, Zhou, Zhang, Jia, Li, and Lc}]{Yang2022}
Yang, D.; Zhou, Y.; Zhang, Z.; Jia, T.; Li, J.; and Lc, R. 2022.
\newblock {AI as an Active Writer: Interaction strategies with generated text
  in human-AI collaborative fiction writing}.
\newblock In \emph{Joint Proceedings of the ACM IUI Workshops 2022}. Helsinki,
  Finland.

\bibitem[{Yuan et~al.(2021)Yuan, Cao, Wang, and Yarosh}]{yuan_tabletop_2021}
Yuan, Y.; Cao, J.; Wang, R.; and Yarosh, S. 2021.
\newblock Tabletop {Games} in the {Age} of {Remote} {Collaboration}: {Design}
  {Opportunities} for a {Socially} {Connected} {Game} {Experience}.
\newblock In \emph{Proceedings of the 2021 {CHI} {Conference} on {Human}
  {Factors} in {Computing} {Systems}}, 1--14. Yokohama Japan: ACM.
\newblock ISBN 978-1-4503-8096-6.

\bibitem[{Zhou et~al.(2023)Zhou, Zhu, Hu, Pujara, Ren, Callison-Burch, Choi,
  and Ammanabrolu}]{Zhou2022}
Zhou, P.; Zhu, A.; Hu, J.; Pujara, J.; Ren, X.; Callison-Burch, C.; Choi, Y.;
  and Ammanabrolu, P. 2023.
\newblock {I} Cast Detect Thoughts: Learning to Converse and Guide with Intents
  and Theory-of-Mind in Dungeons and Dragons.
\newblock In \emph{Proceedings of the 61st Annual Meeting of the Association
  for Computational Linguistics (Volume 1: Long Papers)}, 11136--11155.
  Toronto, Canada: Association for Computational Linguistics.

\bibitem[{Zhu et~al.(2023)Zhu, Aggarwal, Feng, Martin, and
  Callison-Burch}]{FIREBALL}
Zhu, A.; Aggarwal, K.; Feng, A.; Martin, L.; and Callison-Burch, C. 2023.
\newblock {FIREBALL}: A Dataset of Dungeons and Dragons Actual-Play with
  Structured Game State Information.
\newblock In \emph{Proceedings of the 61st Annual Meeting of the Association
  for Computational Linguistics (Volume 1: Long Papers)}, 4171--4193. Toronto,
  Canada: Association for Computational Linguistics.

\bibitem[{Zhu and {D\&D Beyond}(2016)}]{Avrae}
Zhu, A.; and {D\&D Beyond}. 2016.
\newblock Avrae.
\newblock \url{https://avrae.io/}.

\end{thebibliography}

\end{document}